\documentclass[wcp]{jmlr}

\usepackage{amsfonts}
\usepackage{quotes}

\def\Normal{{\mathcal N}}
\def\w{{\boldsymbol \omega}}
\def\F{{\mathbf F}}
\def\x{{\mathbf x}}
\def\z{{\mathbf z}}

\def\yhead{\hat{y}}
\def\yheadN{\hat{y}_{\Normal}}
\def\R{\mathbb{R}}
\def\I{{\mathbf I}}
\def\b1{{\mathbf 1}}

\DeclareMathOperator*{\argmin}{arg\,min}
\def\E{\mathop{{\mathbb E}}}

\usepackage{longtable}

\usepackage{booktabs}

\usepackage{lineno}

\makeatletter
\let\Ginclude@graphics\@org@Ginclude@graphics 
\makeatother

\jmlrvolume{222}
\jmlryear{2023}
\jmlrworkshop{ACML 2023}

\title[Optimal Nonlinearities Improve Generalization Performance of Random Features]{Optimal Nonlinearities Improve \\ Generalization Performance of Random Features}

\author{\Name{Samet Demir} \Email{sdemir20@ku.edu.tr}\\
\addr Machine Learning and Information Processing Group, KUIS AI Center, Ko\c{c} University, Turkey
\AND
\Name{Zafer Do\u{g}an} \Email{zdogan@ku.edu.tr}\\
\addr Machine Learning and Information Processing Group, KUIS AI Center, Ko\c{c} University, Turkey \\
Electrical and Electronics Engineering, Ko\c{c} University, Turkey
}

\editors{Berrin Yan{\i}ko\u{g}lu and Wray Buntine}

\begin{document}

\maketitle

\begin{abstract}
Random feature model with a nonlinear activation function has been shown to perform asymptotically equivalent to a Gaussian model in terms of training and generalization errors. Analysis of the equivalent model reveals an important yet not fully understood role played by the activation function. To address this issue,  we study the "parameters" of the equivalent model to achieve improved generalization performance for a given supervised learning problem. We show that acquired parameters from the Gaussian model enable us to define a set of optimal nonlinearities. We provide two example classes from this set, e.g., second-order polynomial and piecewise linear functions. These functions are optimized to improve generalization performance regardless of the actual form. We experiment with regression and classification problems, including synthetic and real (e.g., CIFAR10) data. Our numerical results validate that the optimized nonlinearities achieve better generalization performance than widely-used nonlinear functions such as ReLU. Furthermore, we illustrate that the proposed nonlinearities also mitigate the so-called double descent phenomenon, which is known as the non-monotonic generalization performance regarding the sample size and the model size. 
\end{abstract}


\maketitle

\section{Introduction}
\label{sec:intro}
We consider a supervised learning problem of fitting a collection of training samples $\{(\mathbf{x}_i, y_i)\}_{i=1}^m$ using the random feature model (RFM) \citep{rahimi2007random}:
\begin{equation}
    \label{eq:feature_model}
    \yhead_{RF} := \w^T\sigma(\F^T \x),
\end{equation}
where $\x \in \R^n$ is an input vector, $\F \in \R^{n \times k}$ is a random feature matrix drawn from some matrix ensembles, $\sigma: \R \to \R$ is an element-wise nonlinear mapping (i.e., activation function), and $\w \in \R^{k}$ is called the weight vector. Note that, in the RFM, the feature matrix $\F$ is fixed after random sampling. As such, one can also view this model as a two-layer network with $k$ hidden neurons where the first layer weights are frozen in the learning process. Hence, only the weight vector $\w$ is learned by solving the following optimization problem:
\begin{equation}
    \label{eq:feature_form}
    \hat{\w}=\underset{\w\in\mathbb{R}^k}{\mbox{argmin}}\frac{1}{m}\sum_{i=1}^m l(y_i, \w^T\sigma(\F^T \x_i)) + \frac{\lambda}{2} ||\w||_2^2,
\end{equation}
where $l:\R^2 \to \R$ is a (convex) loss function, and $\lambda>0$ is a regularization constant. Then, we measure the performance of the learning process via the generalization error defined as
\begin{equation}
    \label{eq:generalization}
    \E_{(\x,y) \sim D} l(y, \hat{\w}^T\sigma(\F^T \x)),
\end{equation}
where $\hat{\w}$ denotes the optimal solution of \eqref{eq:feature_form}, and $D$ refers to the data distribution.

The RFM was initially proposed to approximate kernel methods with linear models \citep{rahimi2007random}. This approach allowed for faster and more scalable computations, making kernel methods feasible for larger datasets. Since then, random features, in general, have been used under different settings \citep{brault2016random,nishio2019random} including multiple kernel learning \citep{bektacs2022fast}. Furthermore, the RFM is also shown to outperform traditional linear models while still being more efficient than kernel methods. Therefore, the RFM has been applied to various problems in machine learning and signal processing \citep{9495136}. 

The RFM has received considerable interest in the last few years mainly due to its simplicity, empirical performance, and connection to overparameterized neural networks \citep{Bach2017, Jacot2018}. Some of that attention has been directed toward characterizing the generalization performance of this model in high-dimensional regimes \citep{mei2022generalization, Ba2020Generalization, mel2022anisotropic}. In this regard, the asymptotic equivalence of the RFM and a Gaussian model have been observed and validated empirically in several papers in the literature \citep{mei2022generalization,montanari2019generalization,gerace2020generalisation}. 

Theoretically, the asymptotic equivalence of the RFM and the Gaussian model has been initially predicted \citep{pmlr-v40-Thrampoulidis15} by using a non-rigorous method from statistical physics (known as the replica method \citep{doi:10.1142/0271}). Recently, these predictions have been verified rigorously, and an underlying universality theorem for the RFM has been proved \citep{hu2022universality}. Furthermore, using the equivalent model, the performance of the RFM in the overparameterized regime is precisely characterized by the Gaussian min-max theorem \citep{dhifallah2020precise}. 

Our work is based on the asymptotically equivalent Gaussian formulation of the RFM \eqref{eq:feature_model} provided as follows:
\begin{align}
    \label{eq:gaussian_model}
    \yheadN &:= \w^T(\mu_0\b1 + \mu_1\F^T\x + \mu_2\z),
\end{align}
where $\b1$ is an all-one vector and  $\z \overset{\mbox{\tiny{i.i.d.}}}{\sim}\Normal(0,\I_k)$ is independent of $\x$. Moreover, for a given $\sigma(\cdot)$, the quantities $\mu_0, \mu_1$, and $\mu_2$, which we call "mapping parameters", are defined as $\mu_0=\E[\sigma(z)], \quad \mu_1 = \E[z\sigma(z)],\mbox{ and } \mu_2 = ({\E[\sigma(z)^2] - \mu_1^2 - \mu_0^2})^{1/2}$ where $z \sim \mathcal{N}(0,1)$. Specifically, training and generalization performances (errors) for \eqref{eq:feature_model} are asymptotically equivalent to those for \eqref{eq:gaussian_model}. This holds for some reasonable feature matrix $\F$, loss function $l(\cdot, \cdot)$, and nonlinearity $\sigma(\cdot)$. In addition, it also requires standard Gaussian inputs and labels that are generated using a typical teacher-student framework, which is a common technique in the theoretical literature \citep{Loureiro_2022,wang2021understanding,cao2022towards}.

In this work, we focus on the equivalent model \eqref{eq:gaussian_model} to study the effects of the nonlinearity in the RFM \eqref{eq:feature_model} to achieve improved generalization performance over a set of learning problems. In the literature, the effects of regularization \citep{nakkiran2020optimal} and data-dependent feature selection \citep{shahrampour2018data} on the generalization performance of RFM have been studied. Furthermore, the effects of different nonlinear activation functions in the RFM have been observed before \citep{dhifallah2020precise, hu2022universality} not to mention a vast literature \citep{agostinelli2015learning, ramachandran2018searching, nwankpa2018activation} on activation functions in general. However, the precise characterization of the nonlinearity over the generalization performance of the RFM is still required. In a concurrent work on "optimal activation functions", a combination of generalization error and sensitivity of the activation functions is minimized under a regression setting with a synthetic data model \citep{wang2023optimal}. On the other hand, we focus on minimizing the generalization error under various settings with synthetic and real data. 

Another line of related work is about the so-called "double descent" phenomenon \citep{Belkin19, pmlr-v80-belkin18a} observed in the generalization performance of the RFM. The phenomenon is characterized by the non-monotonic decrease of the generalization error regarding the model complexity.  In this sense, an optimal choice of the regularization constant $\lambda$ in \eqref{eq:feature_form} has been shown to improve the generalization performance and mitigate the double descent phenomenon \citep{nakkiran2020optimal}. For the RFM, the equivalent formulation \eqref{eq:gaussian_model} suggests that the optimal nonlinearities can be linked to the optimal choice of the regularization constant. 

Overall, the main contributions of this work are two-fold: First, we improve the generalization performance of the RFM by using optimal nonlinearities derived from the mapping parameters of the asymptotically equivalent Gaussian formulation. Second, we show that the proposed nonlinearities also mitigate the double descent phenomenon. 

\section{Optimal Nonlinearities}
\label{sec:approach}
Comparing the RFM and the equivalent formulation, the effect of the nonlinearity in the former only appears as the mapping parameters ($\mu_{0,1,2}$) in the latter. As such, we first replace the RFM with the equivalent model and optimize the mapping parameters to minimize the generalization error. Next, we show that for a given set of optimized mapping parameters (optimal in the sense of generalization performance), it is possible to define a set of optimal nonlinearities that perform asymptotically equivalent. Finally, we provide two example classes of such nonlinear functions to be replaced with $\sigma$ in the RFM to illustrate the performance of the "optimal" nonlinearities. 

\subsection{Assumptions}
\label{sec:assumptions}
Our results are based on the following technical assumptions for the equivalence of the RFM and the Gaussian model \citep{hu2022universality}: 
\begin{enumerate}
    \item The data vectors $\{\mathbf{x}_i\}_{i=1}^{m}$ are drawn independently from $\mathcal{N}(0,\I_n)$.
    \item The number of samples $m$, the input dimension $n$, and the feature dimension $k$ go to infinity with finite ratios $m/n > 0$ and $k/m > 0$.
    \item The labels are generated using a typical teacher-student setup described in the Sec.\ \ref{sec:results}. Furthermore, the unknown signal $\boldsymbol{\xi} \in \R^{n}$ used in the teacher model is independent of the feature matrix $\F$, where $\|\boldsymbol{\xi}\|=\rho$ is known.
    \item The nonlinear function $\sigma(\cdot)$ satisfies the conditions that $\mu_1 = E[z\sigma(z)] > 0$ and $0 < \mu_2 = E[\sigma(z)^2] < \infty $, where $z \sim \mathcal{N}(0, 1)$.
    \item The loss function $l(\cdot,\cdot)$ is a proper convex function in $\mathbb{R}^2$.
    \item The columns of the feature matrix $\F$ are independent and identically drawn from a Gaussian distribution with zero mean and covariance matrix $\frac{1}{n}\mathbf{I}_n$. 
\end{enumerate}

\subsection{Optimizing the Mapping Parameters}
\begin{figure}
    \centering
    \includegraphics[width=0.99\textwidth]{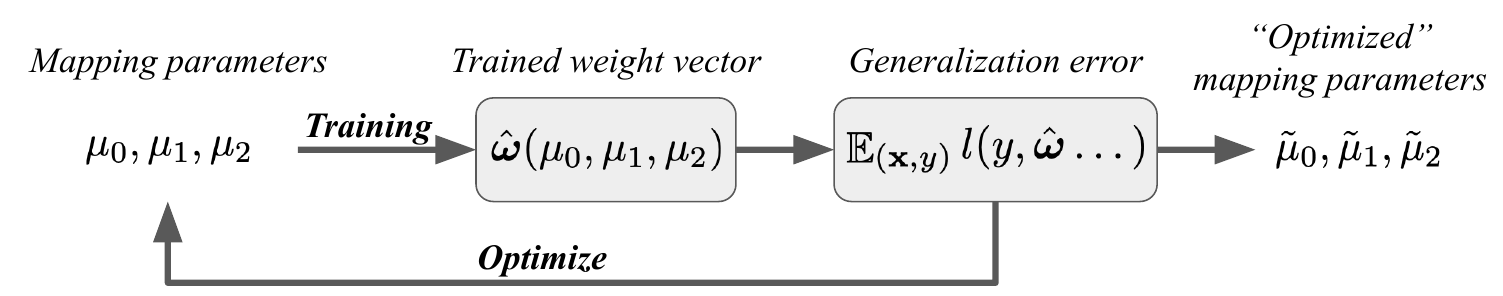}
         \caption{Overview - minimizing the generalization error w.r.t. the mapping parameters}
         \label{fig:optimization_overview}
\end{figure}

\label{sec:optimization}
For a given set of training samples $\{(\mathbf{x}_i,y_i)\}_{i=1}^m$, loss function $l$, feature matrix $\F$, and the regularization constant $\lambda$, we  learn the \textit{optimal mapping parameters}, i.e., $\tilde{\mu}_{0,1,2}$ as illustrated in Fig.~\ref{fig:optimization_overview}. First, we define the trained weight vector $\hat{\w}$ as a function of the mapping parameters:
\begin{equation}
    \label{eq:gaussian_form}
    \hat{\w}(\mu_0, \mu_1, \mu_2) := \underset{\w\in\mathbb{R}^k}{\mbox{argmin}}\frac{1}{m}\sum_{i=1}^m l(y_i, \w^T(\mu_0\b1 + \mu_1\F^T\x_i + \mu_2\z)) + \frac{\lambda}{2} ||\w||_2^2,
\end{equation}
where $\b1$ is an all-one vector and  $\z \overset{\mbox{\tiny{i.i.d.}}}{\sim}\Normal(0,\I_k)$ as in the definition of the equivalent Gaussian model. Then, we can pose the objective of minimizing the generalization error with respect to the mapping parameters:
\begin{equation}
    \label{eq:optimization_obj}
    \tilde{\mu}_{0,1,2} = \argmin_{\mu_0 \in \mathbb{R}, \mu_1, \mu_2 > 0} \quad \E_{(\x,y) \sim D, \z \sim \Normal(0,\I_k)} l(y, \hat{\w}(\mu_0, \mu_1, \mu_2)^T(\mu_0\b1 + \mu_1\F^T\x + \mu_2\z)).
\end{equation}
\eqref{eq:optimization_obj} specifies our general objective. We observe that $\mu_0$ plays the role of the so-called "bias" term. Therefore, we may drop $\mu_0$ from the objective in favor of a bias term $b$:
\begin{equation}
    \label{eq:optimization}
    \tilde{\mu}_1, \tilde{\mu}_2 = \argmin_{\mu_1, \mu_2} \E_{(\x,y) \sim D, \z \sim \Normal(0, \I)} l(y, \hat{\w}(\mu_1, \mu_2)^T(\mu_1\F^T\x + \mu_2\z) + \hat{b}(\mu_1, \mu_2)),
\end{equation}
where $\hat{\w}(\mu_1, \mu_2)$ and $\hat{b}(\mu_1, \mu_2)$ are obtained, similar to \eqref{eq:gaussian_form}, as follows:
\begin{equation}
    \label{eq:w_optimization}
    \hat{\w}(\mu_1, \mu_2), \hat{b}(\mu_1, \mu_2) := \argmin_{\w, b}  \frac{1}{m}\sum_{i=1}^m l(y_i, \w^T(\mu_1\F^T\x_i + \mu_2\z_i) + b) + \frac{\lambda}{2} ||\w||_2^2,
\end{equation}
where $\z \overset{\mbox{\tiny{i.i.d.}}}{\sim}\Normal(0,\I_k)$, $\z_i \overset{\mbox{\tiny{i.i.d.}}}{\sim}\Normal(0,\I_k)$ for $i \in \{1,\dots,m\}$ and $\tilde{\mu}_0 = \hat{b}(\tilde{\mu}_1, \tilde{\mu}_2)/ \hat{\w}(\tilde{\mu}_1, \tilde{\mu}_2)^T\b1$. 
For the experimental results, \eqref{eq:optimization} is solved using grid search, and we use the closed-form solution for \eqref{eq:w_optimization} whenever it is available (depending on the loss). Note that for each feature matrix $\F$, \eqref{eq:optimization} is solved again. Therefore, the optimal mapping parameters are specific to the feature matrix, dataset, and loss function. The complete algorithm for the optimization is provided in Sec.~\ref{sec:algo}. 

\subsection{Mapping Parameters to Nonlinear Mappings}
\label{sec:sigma_generation}

Given mapping parameters ($\mu_{0,1,2}$), we next define a set of nonlinear mappings:
\begin{align}
    \label{eq:function_set}
    \mathcal{F}_{\mu_{0,1,2}} = \left\{\sigma_f: \R \to \R \middle|
    \begin{array}{l}
    \mu_0 = \E_z[\sigma_f(z)],\\
    \mu_1 = \E_z[z\sigma_f(z)],\\
    \mu_2 = \sqrt{\E_z[\sigma_f(z)^2] - \mu_1^2 - \mu_0^2}
    \end{array} \right\},
\end{align}
where $z \sim \Normal(0,1)$. Note that constraints in \eqref{eq:function_set} do not determine a unique mapping without further assumptions on $\sigma_f$. Instead, the constraints specify a set of performance-wise equivalent mappings. Moreover, any linear function cannot satisfy these constraints when $\mu_2\neq0$. Hence, \eqref{eq:function_set} defines a set of nonlinear mappings.

Here, we provide two example classes of functions that enable us to uniquely determine the nonlinear mappings assuming specific function families. First, we consider a naive second-order polynomial with three coefficients to be determined. Using the optimal mapping parameters from \eqref{eq:optimization} and the set of equations in \eqref{eq:function_set}, we obtain the following form:
\begin{equation}
    \sigma_{polynomial}(z) = \tilde{\mu}_0 + \tilde{\mu}_1 z + \frac{\tilde{\mu}_2}{\sqrt{2}} \left(z^2 - 1\right),
\end{equation}
which is equivalent to the orthonormal Hermite polynomial expansion with the three mapping parameters.

Second, we consider a nonlinear mapping in piecewise linear form as a generalization of (leaky) ReLU:
\begin{equation}
    \label{eq:leakyReLu}
    \sigma_{piecewise}(z) = \begin{cases}
    az + c,& \text{if } z\geq 0,\\
    bz + c,& \text{otherwise},
\end{cases}
\end{equation}
where the three parameters $a,b,c$ are to be determined. By solving \eqref{eq:function_set}, we obtain the parameters as $a = \tilde{\mu}_1 + \sqrt{\frac{\pi}{\pi - 2}} \tilde{\mu}_2$, $b = \tilde{\mu}_1 - \sqrt{\frac{\pi}{\pi - 2}} \tilde{\mu}_2$, and $c = \tilde{\mu}_0 - \sqrt{\frac{2}{\pi - 2}} \tilde{\mu}_2$.

The proposed nonlinearities can be directly used in the RFM, providing improved generalization performance by definition. Unlike widely used nonlinear activation functions such as ReLU, and Softplus, the proposed ones are "task-optimized" in the sense that it takes the data, the loss, and the regularization into account. As such, the results can be extended to various other applications for better generalization performance.

Before moving to the results, we want to mention a few points about the "optimal" nonlinearities. Regardless of their form, they all provide asymptotically equivalent yet improved generalization errors. However, the optimal mapping parameters reveal more information on optimal nonlinearity. For example, a non-zero $\tilde{\mu}_0$ states that the optimal nonlinearity cannot be an odd function, although this assumption is often used in the simplification of the theoretical development of the universality laws for learning with the RFM \citep{hu2022universality}. $\tilde{\mu}_1$ plays an important role controlling the scale of $\hat{\w}$. Hence, the optimal nonlinearities remove the need for tuning optimal $\lambda$. $\tilde{\mu}_2$ controls the noise level of the RFM. Furthermore, we can show that \eqref{eq:leakyReLu} can be reduced to (a scaled-version of) widely-used ReLU only if the optimal mapping parameters satisfy a special relationship such that $\tilde{\mu}_0^2=\tilde{\mu}_1^2 (\frac{\pi}{2})=\tilde{\mu}_2^2(\frac{2}{\pi-2})$. It suggests that ReLU will naturally arise as the optimal nonlinearity for a specific set of learning problems depending on the set of training samples, loss function, $\F$, and $\lambda$.

\begin{figure}
    \centering
    \includegraphics[width=0.76\textwidth]{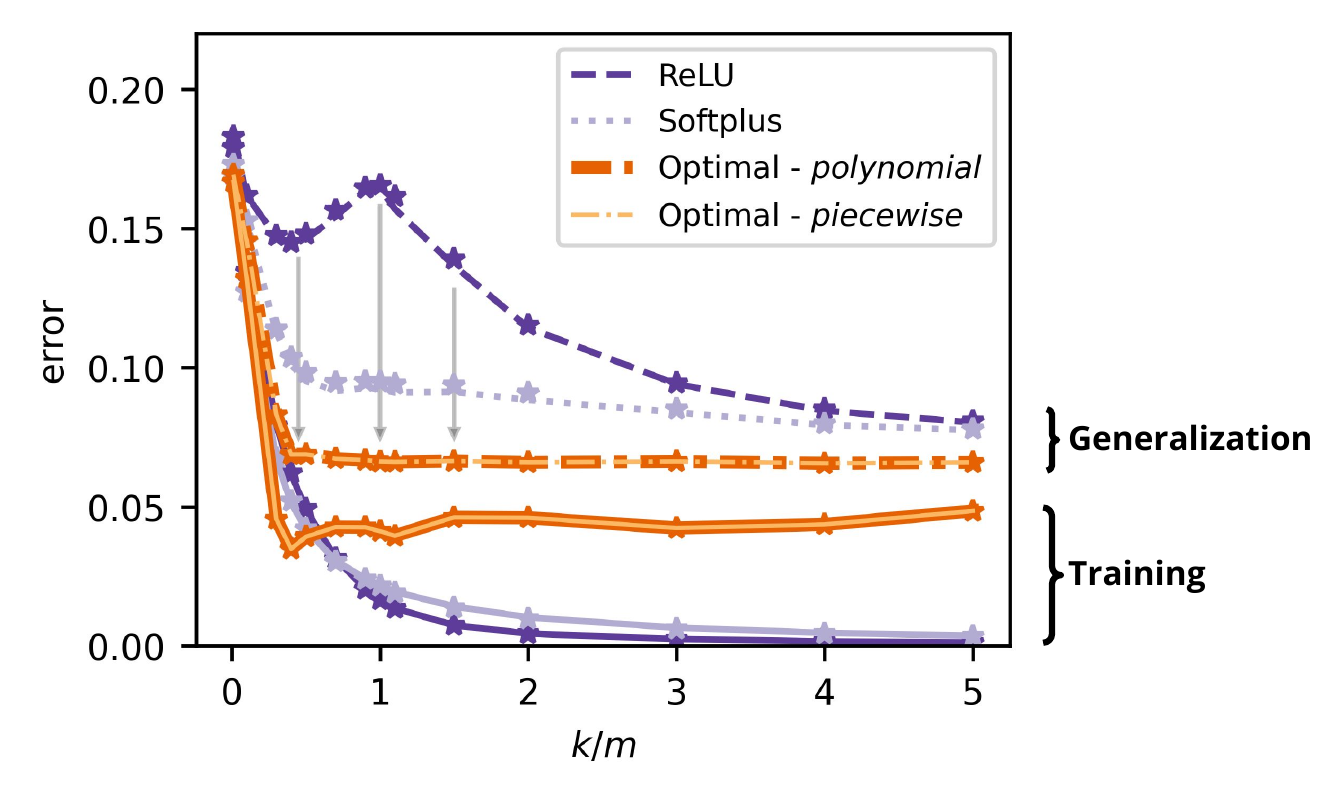}
    \caption{Regression (Sec.~\ref{sec:regression}) - training (solid lines) and generalization (dashed lines) errors are provided for the RFM and the equivalent Gaussian model with ReLU, Softplus, and the proposed optimal nonlinearities. $\star$ denotes the error for the equivalent Gaussian model \eqref{eq:gaussian_model}. The downward arrows show that the generalization error has been improved across the full range of the model complexities we consider. The numerical results are averaged over 50.}
    \label{fig:regression}
\end{figure}

\section{Results}
\label{sec:results}
In this section, we provide experimental results showing that generalization performance is improved when the proposed set of nonlinearities is used in the RFM. To illustrate our results, we consider two applications: nonlinear regression and binary classification problems with different loss functions on synthetic data. We also consider binary classification on CIFAR10 \citep{krizhevsky2009learning} and Tiny ImageNet \citep{wu2017tiny, deng2009imagenet} datasets. All results present the corresponding training and generalization performances of the underlying the RFM with specific choices of the nonlinear activation functions and the proposed optimal nonlinearities as a function of model complexity $(k/m)$. 

\subsection{Experimental Setup}
In this section, we provide the full details of our experimental setup. Specifically, we discuss the optimization algorithm, the data model, the feature matrix, the dimensions, and the regularization constant, which are essential components of the experimental design.

\subsubsection{The Optimization Algorithm}
\label{sec:algo}
To optimize the mapping parameters, we use grid search, which is a widely used technique for hyperparameter optimization in machine learning \citep{bergstra2012random}. We have two reasons to use grid search: 1) our optimization objective is observed to be nonconvex, which limits the application of gradient-based optimization techniques; 2) the number of parameters to search for is two ($\mu_1, \mu_2$), which makes the grid search computationally cheap. One might also use more advanced approaches, such as randomized search, Bayesian optimization, or evolutionary algorithms \citep{bergstra2011algorithms}. However, grid search is simple and efficient enough to show our claims in this paper. The complete algorithm is provided in Algorithm~\ref{alg:optimization}. We first calculate the "bias term" from the labels and initialize the variables for the grid search. For each $(\mu_1, \mu_2)$ pair in the search grid, we train the weight vector $\hat{\w}$ and calculate generalization error with $\hat{\w}$. If there is an improvement in the generalization error, we update the parameters. Finally, $\tilde{\mu}_0$ is calculated. Note that there is a difference between synthetic and real data experiments. We use the error calculated on the validation set instead of the generalization error for the real data, while the test set is reserved for calculating the generalization error.

\begin{algorithm2e}[tbh]
\caption{The algorithm for the optimization}\label{alg:optimization}
\KwData{$\{(\mathbf{x}_i,y_i)\}_{i=1}^m, SearchGrid$}
\KwResult{$\tilde{\mu}_0, \tilde{\mu}_1, \tilde{\mu}_2$}
$\hat{b} \gets \frac{1}{m}\sum_{i=1}^m y_i$\;
$\tilde{\mu}_1 \gets 1$\;
$\tilde{\mu}_2 \gets 0$\;
$\hat{\w}_{best} \gets \b1$\;
$best\_error \gets \infty$\;
\For{$(\mu_1, \mu_2)$ in SearchGrid}{
  $\hat{\w} \gets \underset{\w\in\mathbb{R}^k}{\mbox{argmin}}  \frac{1}{m}\sum_{i=1}^m l(y_i, \w^T(\mu_1\F^T\x_i + \mu_2\z_i) + \hat{b}) + \frac{\lambda}{2} ||\w||_2^2$ \tcp*{Eq.~\eqref{eq:w_optimization}}
  \eIf{Synthetic Data}{
    $error \gets \underset{(\x,y) \sim D, \z \sim \Normal(0, \I)}{\E} l(y, \hat{\w}^T(\mu_1\F^T\x + \mu_2\z) + \hat{b})$ \tcp*{Eq.~\eqref{eq:optimization}}
  }{
    $error \gets \textit{Calculate Error on the Validation Set}$ \tcp*{Eq.~\eqref{eq:optimization} on real data}
  }
  \If{$error < best\_error$}{
    $\tilde{\mu}_1 \gets \mu_1$\;
    $\tilde{\mu}_2 \gets \mu_2$\;
    $\hat{\w}_{best} \gets \hat{\w}$\;
    $best\_error \gets error$\;
  }
}
$\tilde{\mu}_0 \gets \hat{b} / \hat{\w}_{best} \b1$\;
\end{algorithm2e}

\subsubsection{The Data model}
We consider the classical teacher-student framework \citep{Loureiro_2022} for the numerical simulations. Specifically, we assume the following data generation model:
\begin{equation}
    \label{eq:data}
    \x_i \sim \Normal(0, \I_n), \, y_i = \psi(\boldsymbol{\xi}^T \boldsymbol{\x}_i) + \Delta \epsilon_i, \forall i\in\{1,\cdots,m\},
\end{equation}
where $\x_i \in \R^n$ and $y_i \in \R$ are input-output pairs, $\boldsymbol{\xi} \in \R^n$ is an unknown fixed vector, $\epsilon_i \sim \Normal(0,1)$, $\Delta \geq 0$ is a constant controlling the noise level, and $\psi:\R \to \R$ is a nonlinear function. Furthermore, we use a fixed $\boldsymbol{\xi}$ with unit norm that is sampled from $\Normal(0, (1/n)\I_n)$ for the experiments. 

\subsubsection{The Feature Matrix and The Dimensions}
Without loss of generality, $k$ columns of the feature matrix $\F$ are drawn independently from $\Normal(0, (1/n)\I_n)$ in our experiments. Note that results can be extended to alternative feature matrix $\F$ that satisfy the regularity assumptions introduced in \citep{hu2022universality}. The number of samples $m$, the feature dimension $k$, and the input dimension $n$ are set to satisfy the assumptions in Sec.\ \ref{sec:assumptions}.

\subsubsection{The Regularization Constant}
The choice of the regularization constant $\lambda$ in \eqref{eq:feature_form} is known to play a key role in generalization performance, and the optimal $\lambda$ has been shown to mitigate the double descent behavior \citep{nakkiran2020optimal}. Essentially, the regularization constant controls the scale of the weight vector $\w$. In the equivalent Gaussian model  \eqref{eq:gaussian_model}, we can achieve the same control over the scale of $\w$ by scaling the mapping parameters ($\mu_{0,1,2}$). This observation has two direct consequences. First, we do not need to tune the regularization constant $\lambda$ when using the optimal nonlinear mappings. Second, similar to the effect of optimal regularization \citep{nakkiran2020optimal}, we show that the optimized nonlinear mappings also mitigate the double descent phenomenon.

\subsection{Regression}
\label{sec:regression}
\begin{figure}
     \centering
     \subfigure[Classification (Sec.~\ref{sec:classification})]{
     \includegraphics[width=0.45\textwidth]{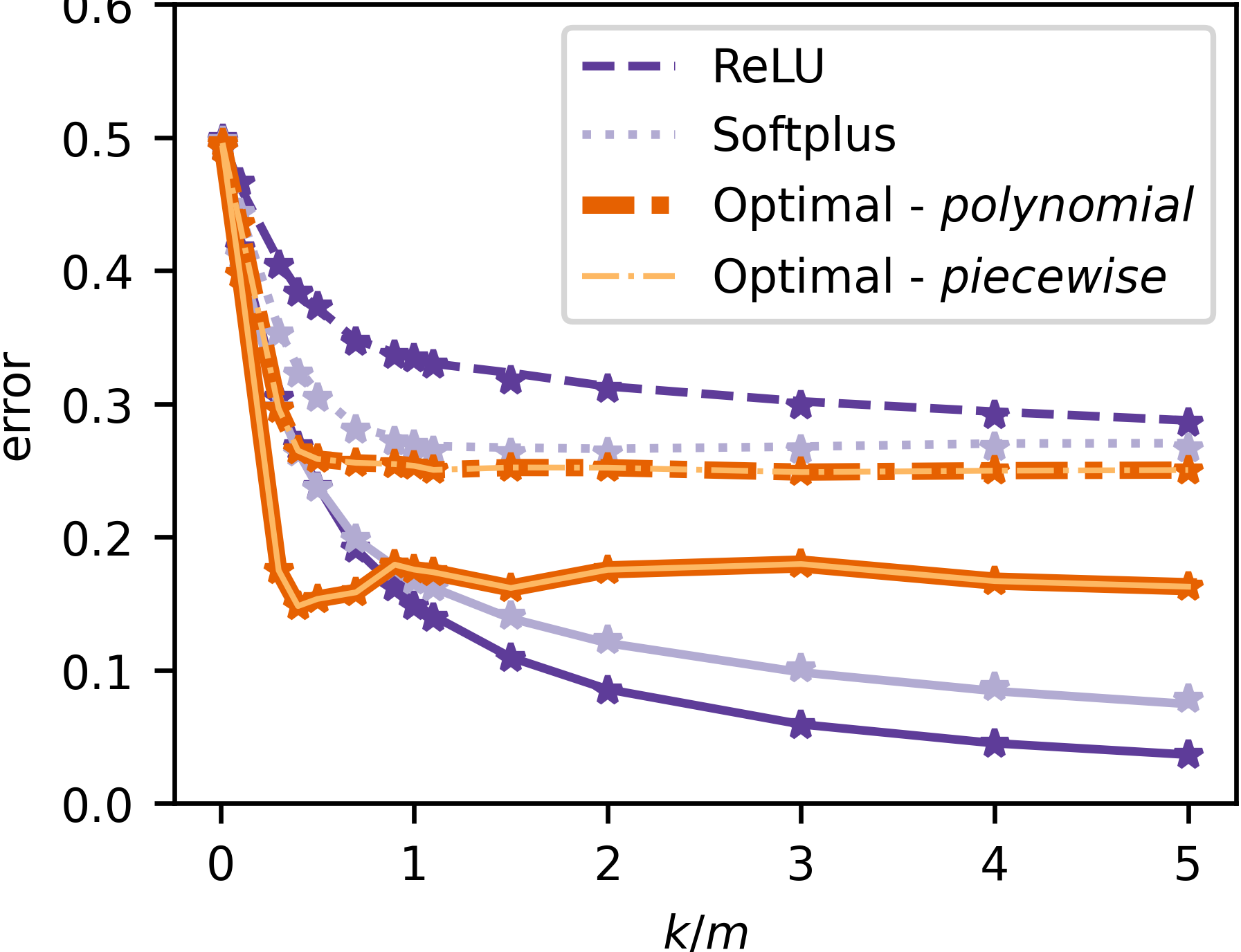}
         \label{fig:classification}
     }
     \hfill
     \subfigure[Classification with smaller $\lambda$]{
     \includegraphics[width=0.45\textwidth]{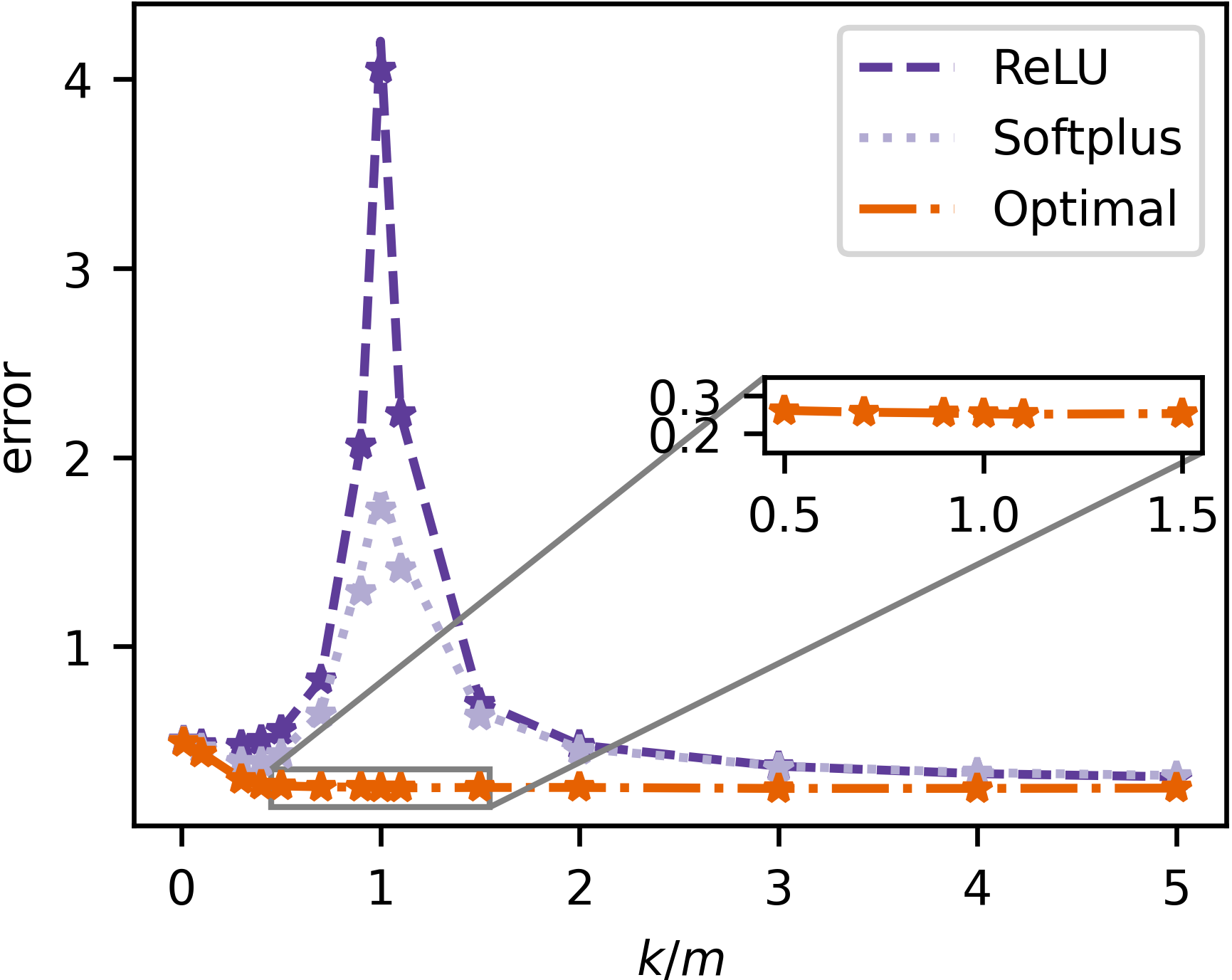}
    \label{fig:double_descent}
     }
        \caption{Numerical simulations - training (solid lines) and generalization (dashed lines) errors are provided for the RFM and the equivalent Gaussian model with ReLU, Softplus, and the proposed optimal nonlinearities. $\star$ denotes the error for the equivalent Gaussian model \eqref{eq:gaussian_model}. The numerical results are averaged over 50.}
        \label{fig:classification_dd}
\end{figure}

Consider a nonlinear regression problem where the labels $\{y_i\}_{i=1}^m$ are generated according to \eqref{eq:data} with $\psi(z) = \max(z,0)$, and $\Delta=0.05$. Moreover, we assume that the loss function is the squared loss: $ l(y, \yhead) = \frac{1}{2} (y - \yhead)^2$.
Note that this is a standard setting for typical regression problems. Also, we set $n=400$, $m=1200$, and $\lambda=10^{-2}$. Fig.~\ref{fig:regression} shows the training and the generalization performance of learning with the RFM and the equivalent Gaussian model for the regression problem (similarly Fig.~\ref{fig:classification} for the classification problem below). The numerical results show that for a given activation function (ReLU or Softplus), the two model's errors are in excellent agreement, aligned with the previous results in the literature \citep{dhifallah2020precise, hu2022universality}. 
Note that it was also shown that the training and the generalization errors of the Gaussian formulation converge in probability to deterministic limiting functions as the dimensions $m,n,k$ tend to infinity. Moreover, the parameters of the limiting functions can be determined by some fixed point equations \citep{dhifallah2020precise}. For brevity, we omit discussions of the analytical predictions. Instead, we focus on improving the generalization performance of the RFM using the proposed optimal nonlinearities. 
First, the two proposed nonlinear mappings, i.e.,  $\sigma_{polynomial}$, and  $\sigma_{piecewise}$, provide equivalent performance on training and generalization errors by definition. Second, they outperform Softplus and ReLU activation functions in the sense that they provide a lower generalization error for the range of model complexity we consider. This result confirms the key role played by the activation function and also validates the improved generalization performance by using optimal nonlinearities for the RFM.

\subsection{Classification}
\label{sec:classification}
As a second example, we consider a binary classification problem where the labels $\{y_i\}_{i=1}^m$ are generated according to \eqref{eq:data} with $\psi(z) = \mbox{sign}(z)$ and $\Delta = 0$. Here, we use $n=200$, $m=600$ and $\lambda=10^{-1}$ for the experiments. In Fig.~\ref{fig:classification}, the training and generalization errors for the RFM and the equivalent Gaussian model are plotted for the classification problem with the squared loss. Similar to the regression case, the proposed nonlinear mappings ($\sigma_{polynomial}$, and  $\sigma_{piecewise}$) provide equivalent performance while they provide improved generalization performance compared to Softplus and ReLU.

\subsubsection{Double Descent Phenomenon}
The generalization error is known to follow a U-shaped curve for small model complexity until reaching a peak known as the interpolation threshold. After the peak, the generalization error decreases monotonically as a function of the model complexity. This behavior is known as the "double descent" phenomenon \citep{Belkin19, pmlr-v80-belkin18a}.
To highlight it, we consider binary classification with the squared loss and set $\lambda=10^{-4}$, $n=400$, $m=1200$. In Fig.~\ref{fig:double_descent}, while ReLU and Softplus experience a steep double descent with a peak at $k/m=1$, optimal nonlinear mappings lead to a monotonically decreasing generalization error. Specifically, comparing the generalization performance in Fig.~\ref{fig:classification} and Fig.~\ref{fig:double_descent} for ReLU or Softplus, the results confirm that optimal regularization plays a key role in mitigating the double descent phenomenon, matching the results stated in \citep{nakkiran2020optimal}. However, regardless of the choice of regularization constant, the optimized nonlinear mappings always achieve a monotonically decreasing generalization error. 

\begin{figure}
    \centering
    \includegraphics[width=0.5\textwidth]{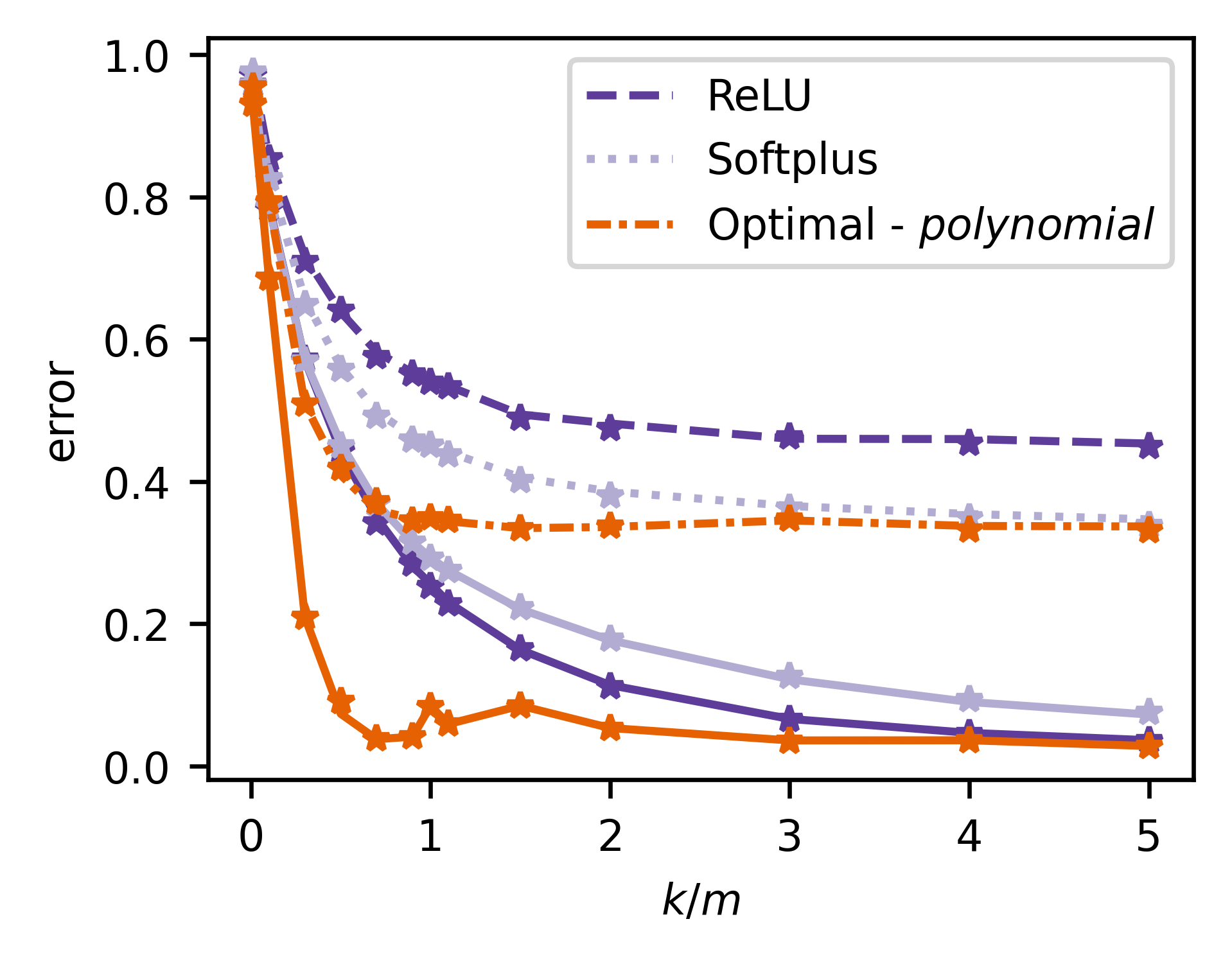}
    \caption{Classification with hinge loss - training (solid lines) and generalization (dashed lines) errors are provided for the RFM and the equivalent Gaussian model with ReLU, Softplus, and the proposed optimal nonlinearities. $\star$ denotes the error for the equivalent Gaussian model \eqref{eq:gaussian_model}. The numerical results are averaged over 20.}
    \label{fig:classification_hinge}
\end{figure}

\subsubsection{Impact of Loss Functions:}
To see the impact of different loss functions, we next consider the hinge loss: $l(y, \yhead) = \max(1-y\yhead, 0)$. Fig.~\ref{fig:classification_hinge} shows the training and generalization errors for the classification case with the hinge loss and $\lambda=10^{-1}$. Note that we use a smaller search grid for this plot since there is no closed-form solution for $\hat{\w}$ in this case. The optimal nonlinearity provides lower training and generalization errors compared to Softplus and ReLU. Unlike the results in Fig.~\ref{fig:classification}, the optimal nonlinearity achieves improved generalization performance without deteriorating the training performance. This result suggests that an optimized loss function can improve the two errors without worsening the other.

\begin{figure}
     \centering

     \subfigure[CIFAR10]{
     \includegraphics[width=0.45\textwidth]{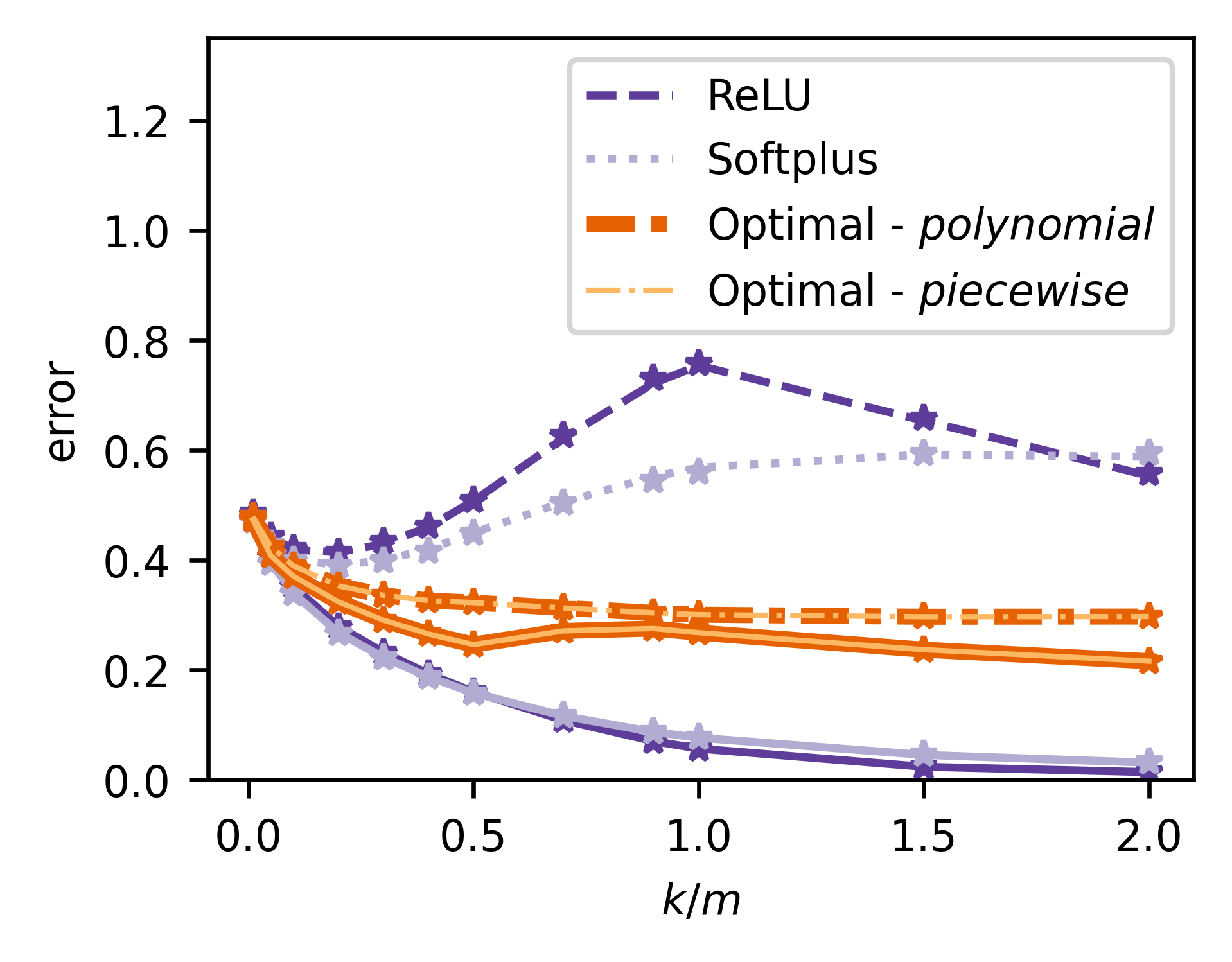}
    \label{fig:cifar10}
     }
     \hfill
     \subfigure[Tiny ImageNet]{
     \includegraphics[width=0.45\textwidth]{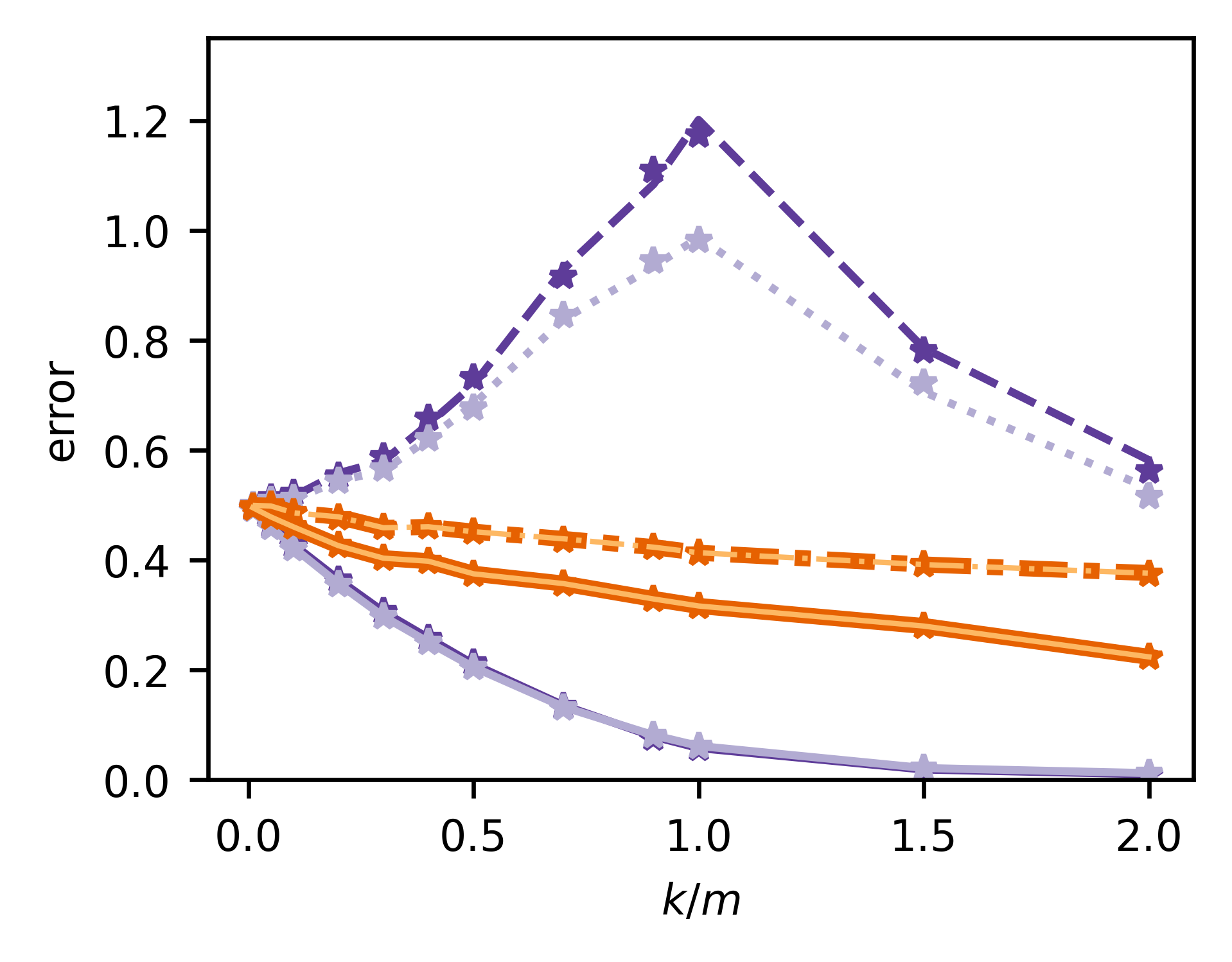}
         \label{fig:imnet}
     }
        \caption{Binary image classification - training (solid lines) and generalization (dashed lines) errors are illustrated for the RFM and the equivalent Gaussian model (denoted by $\star$). The average of 50 Monte Carlo trials is plotted.}
        \label{fig:real_data}
\end{figure}

\subsection{Real Image Classification}
The equivalence of the RFM and the Gaussian model is only valid under assumptions in \ref{sec:assumptions}. For a real dataset, such assumptions will not hold in general. However, these requirements can be satisfied (at least partially) after proper preprocessing.  In this setting, we propose to use a pretrained normalizing flow model \citep{rezende2015variational} (e.g., RealNVP \citep{dinh2016density}) for the preprocessing. Normalizing Flow models are invertible generative models that consist of a sequence of invertible nonlinear transformations. Similar to the popular generative adversarial networks (GAN), the latent space has the standard Gaussian distribution. By taking advantage of the invertibility and the Gaussian latent space, we map real images to Gaussian-distributed latent space. We use real labels in contrast to the teacher-student framework. Another important point regarding our real data experiments is that we optimize the mapping parameters on a validation set and use the test set to calculate the generalization error. 

To extend our results to real data, we consider binary image classification (the first class vs. the second class) on CIFAR10 and Tiny ImageNet. Specifically, for CIFAR10, we pick 2000 samples and 500 samples from each class for the training set and validation sets, respectively. We use the complete test set (1000 samples for each class). For Tiny ImageNet, we use the complete training (500 samples for each class) and validation (50 samples for each class) sets. The validation set is used as the test set since the available test set is not labeled. Furthermore, the training set is split into training (400 samples for each class) and validation (100 samples for each class) sets. We convert the labels such that $y_i \in \{-1,1\}$. In order to achieve the asymptotic equivalence of the RFM and the Gaussian model on real data, the inputs are preprocessed with a pretrained RealNVP model to map them to the latent space, which has Gaussian distribution (see Appendix A for details). The RealNVP model trained in \citep{goldt2022gaussian} is used for CIFAR10, while we trained another RealNVP model for Tiny ImageNet. We use the squared loss and set $\lambda = 10^{-2}$. Fig.~\ref{fig:real_data} illustrates the results on the real data. The results validate that the proposed optimal nonlinearities achieve improved generalization performance and also mitigate the double descent phenomenon. Furthermore, we provide accuracy plots for the same setting in Appendix D, which suggests our results on the error metrics can be extended to other metrics such as accuracy. 

Note that we observe an improved generalization performance with all real image classification problems we consider, provided that a proper normalizing flow model is available. As far as we know, this proof-of-concept result is the first to demonstrate the practical usage of equivalent Gaussian models of RFM to improve the generalization performance with real data. However, such a normalizing flow model might not always be available in practice, which is a current limitation of this work, and alternative solutions are left for future work.

\section{Conclusion} 
In this work, we studied the role played by the nonlinear activation function in the random feature model, which has been shown to perform asymptotically equivalent to a linear Gaussian model. By studying the mapping parameters of the equivalent model, we define a set of optimal nonlinearities that provide equivalent yet improved generalization performance. The proposed nonlinearities achieve better generalization performance than widely-used nonlinear functions. Additionally, we show that the proposed nonlinearities also achieve a monotonic decrease in generalization error. The results are valid for a large family of feature matrices, activation functions, and convex loss functions. Experimental results on classification and regression problems validate the effectiveness of the proposed optimal nonlinearities in achieving better generalization performance.

\acks{We acknowledge that this work was supported in part by TUBITAK 2232 International Fellowship for Outstanding Researchers Award (No. 118C337) and an AI Fellowship provided by Koç University \& İş Bank Artificial Intelligence (KUIS AI) Research Center.}

\bibliography{refs}

\appendix

\section{Preprocessing with Normalizing Flow}
\begin{figure}[tbh]
    \centering
    \includegraphics[width=0.8\textwidth]{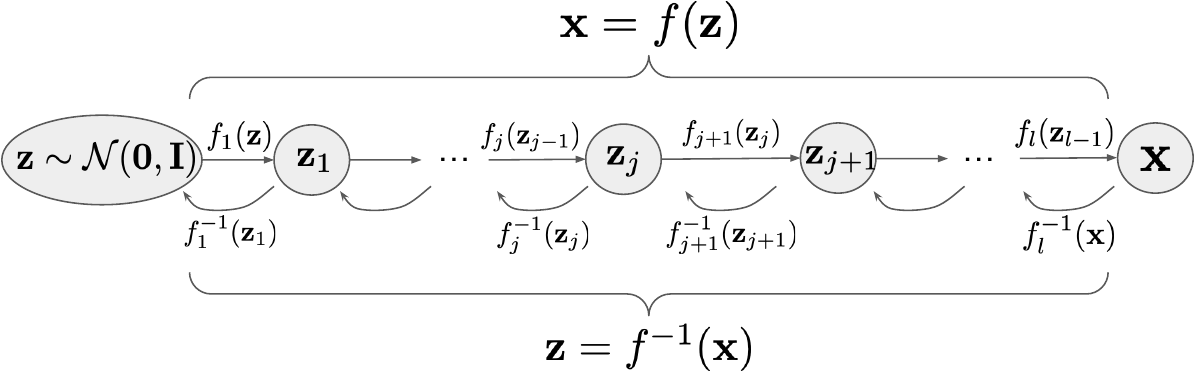}
    \caption{Normalizing flows}
    \label{fig:nf}
\end{figure}

Normalizing Flow is a likelihood-based generative model that models the data distribution by applying a series of invertible (possibly nonlinear) transformations to multivariate standard Gaussian distribution \citep{rezende2015variational}. A Normalizing Flow models consist of a sequence of invertible mappings as illustrated in Fig.~\ref{fig:nf}. Overall, it learns an invertable mapping $f: \mathbb{R}^n \to \mathbb{R}^n$ with which a data sample can be generated by $\x= f(\z)$ where $\z \sim \mathcal{N}(0, \I_n)$. We are interested in the inverse mapping $f^{-1}(\x)$ since our goal is to map real data to Gaussian distribution so that the Gaussian equivalence holds. We process the samples of the dataset as $\hat{\x}_i = f^{-1}(\x_i)$. Then, the collection of preprocessed samples $\{(\hat{\x}_i, y_i)\}$ are used to train and test the random feature model.

\section{CIFAR10 Experiments}
We use a subset of the CIFAR10 dataset \citep{krizhevsky2009learning}. The CIFAR10 dataset includes colored 32x32 images with labels for ten classes. For the training set, we pick 2000 samples from each of the first two classes to form a binary classification problem (airplane vs. automobile). Also, we convert the labels such that $y_i \in \{-1,1\}$. For the test set, we use all the test samples for these two classes. Then, we consider a binary classification problem with the squared loss on the 4000 training, 1000 validation, and 2000 test samples. Furthermore, we use RealNVP \citep{dinh2016density} model that is pretrained and used in \citep{goldt2022gaussian} as the Normalizing Flow model. The model is pretrained on 46000 samples (including all ten classes) from the CIFAR10 dataset, and it is reported to achieve 3.5 bits/dim on a validation set of 4000 samples \citep{goldt2022gaussian}. 

\section{Tiny ImageNet Experiments} 
We also experiment with a small version of ImageNet \citep{wu2017tiny, deng2009imagenet} dataset called "Tiny ImageNet". This dataset consists of 64x64 colorful images. Similar to our previous settings, we focus on the binary classification problem (goldfish vs. fire salamander) with squared loss when the data is inverted with RealNVP. In this case, the training set has 800 samples; the validation set contains 200 samples, while the test set includes 100. We trained the RealNVP model on the ImageNet dataset using the same code used for CIFAR10 and used it in our experiments. The rest
of the setting is the same as CIFAR10 experiments.

\section{Accuracy Plots}

\begin{figure}
     \centering

     \subfigure[CIFAR10]{
     \includegraphics[width=0.54\textwidth]{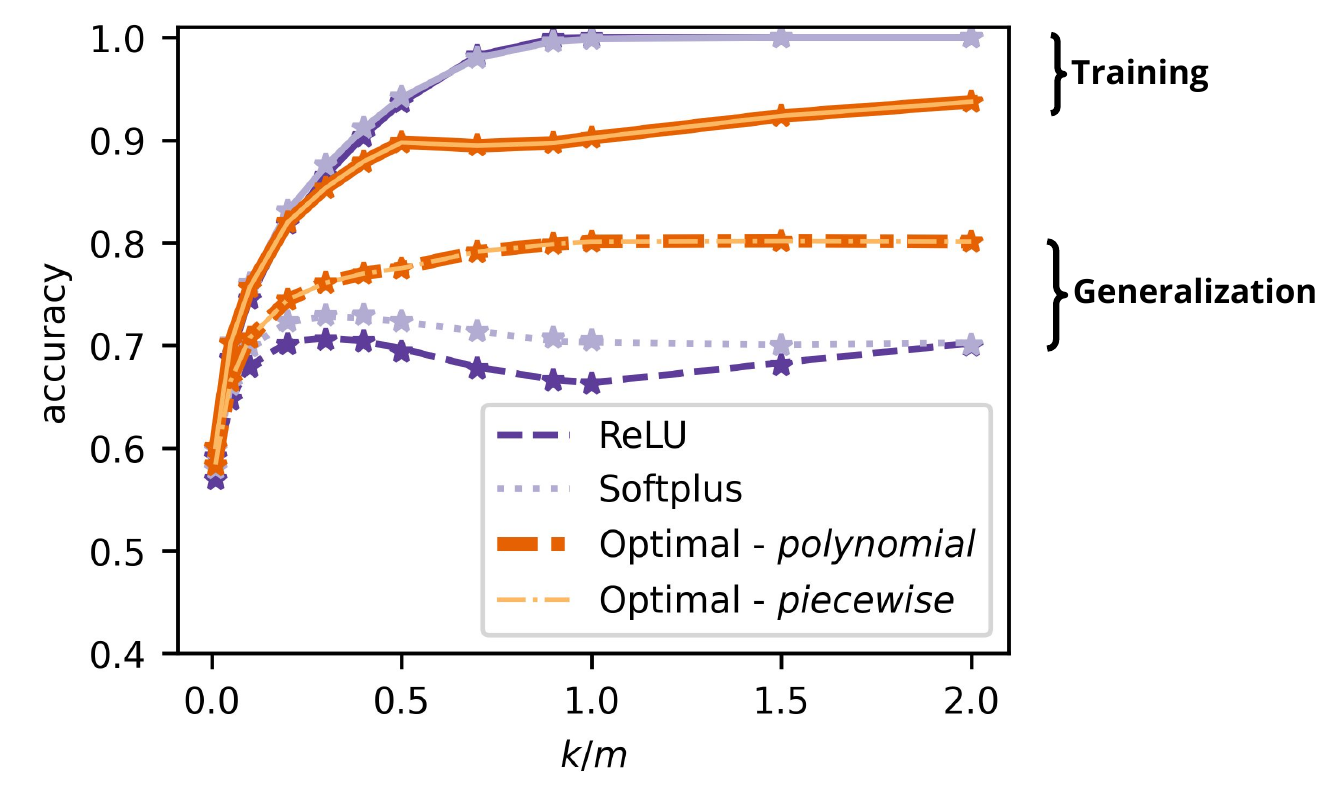}
    \label{fig:cifar10_acc}
     }
     \hfill
     \subfigure[Tiny ImageNet]{
     \includegraphics[width=0.41\textwidth]{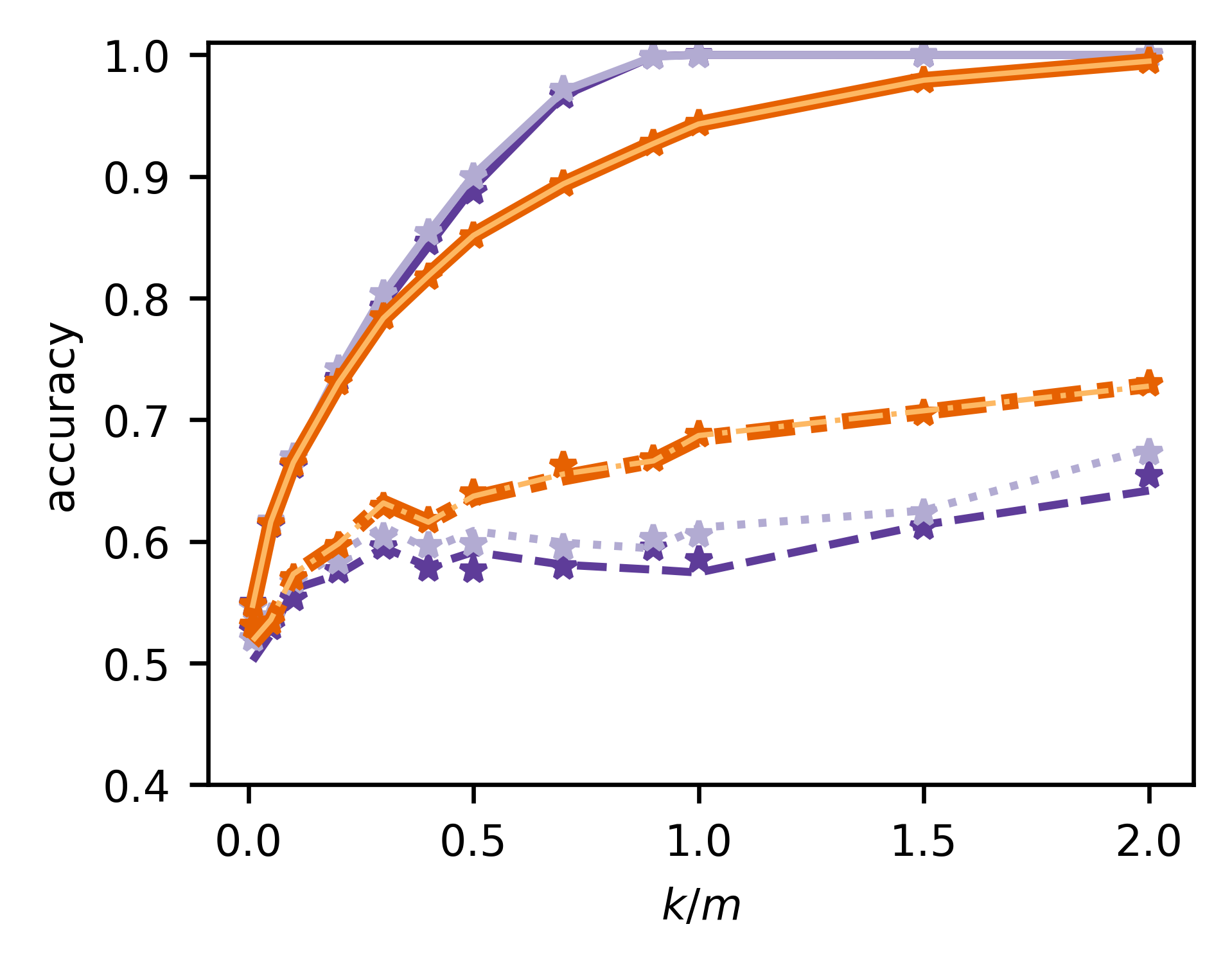}
         \label{fig:imnet_acc}
     }
        \caption{Accuracy plots for binary image classification -  training (solid lines) and generalization (dashed lines) accuracy values are illustrated for the RFM and the equivalent Gaussian model (denoted by $\star$). The same setting as Fig.~\ref{fig:real_data} is used. The average of 50 Monte Carlo runs is plotted.}
        \label{fig:real_acc}
\end{figure}

In Fig.~\ref{fig:real_acc}, we provide an additional plot for the accuracy of the random feature model for different model complexity ($k/m$) values under the same setting as Fig.~\ref{fig:real_data} (in the main text). We observe two points. First, there is an agreement between the accuracy values of the RFM and those of the equivalent Gaussian model. Second, the proposed nonlinearity significantly outperforms the other nonlinearities for the complete range of $k/m$ values we consider. Note that there is an upside-down version of the double descent phenomenon (especially for ReLU) in Fig.~\ref{fig:cifar10_acc}. Although Fig.~\ref{fig:cifar10_acc} looks smooth, we observe fluctuations in Fig.~\ref{fig:imnet_acc}, which is related to the fact that there exist a small number of samples for Tiny ImageNet experiments.

\end{document}